\documentclass[utf8]{frontiersSCNS} % for Science, Engineering and Humanities and Social Sciences articles
\usepackage{url,hyperref,lineno,microtype}
\usepackage[onehalfspacing]{setspace}
\usepackage{cleveref}
\usepackage{microtype,subcaption}
\usepackage[onehalfspacing]{setspace}
\usepackage{tikz}
\usepackage{stfloats}
\usepackage{multicol}
\usepackage{multirow}
\usepackage{rotating}
\usepackage{mathtools}
\usepackage{soul}
\usepackage{framed}
\graphicspath{{images/}{jpg/}}
\usepackage{mathptmx} 
\usepackage{times}
\usepackage{amsmath} 
\usepackage{amssymb}
\usepackage[]{xcolor}
\usepackage[normalem]{ulem}
\usepackage{caption}
\usepackage{siunitx}
\usepackage{graphicx}
\usepackage{listings}
\usepackage{xcolor}

\usepackage[labelformat=parens,labelsep=quad, skip=3pt]{caption}
\def\keyFont{\fontsize{8}{11}\helveticabold }

\def\firstAuthorLast{Shafiee {et~al.}} %use et al only if is more than 1 author
\def\Authors{Milad Shafiee Ashtiani\,$^{\ddagger}$, Alborz Aghamaleki Sarvestani\,$^{\ddagger}$ and Alexander Badri-Spr\"owitz\,$^{*}$}

\def\Address{Dynamic Locomotion Group, Max Planck Institute for Intelligent Systems, Stuttgart, Germany\\
$^{\ddagger}$Equal Contribution}

\def\corrAuthor{Alexander Badri-Spr\"owitz}
\def\corrEmail{sprowitz@is.mpg.de}

\definecolor{codegreen}{rgb}{0,0.6,0}
\definecolor{codegray}{rgb}{0.5,0.5,0.5}
\definecolor{codepurple}{rgb}{0.58,0,0.82}
\definecolor{backcolour}{rgb}{0.95,0.95,0.92}

\lstdefinestyle{mystyle}{
  backgroundcolor=\color{backcolour},   commentstyle=\color{codegreen},
  keywordstyle=\color{magenta},
  numberstyle=\tiny\color{codegray},
  stringstyle=\color{codepurple},
  basicstyle=\ttfamily\footnotesize,
  breakatwhitespace=false,         
  breaklines=true,                 
  captionpos=b,                    
  keepspaces=true,                 
  numbers=left,                    
  numbersep=5pt,                  
  showspaces=false,                
  showstringspaces=false,
  showtabs=false,                  
  tabsize=2
}

\begin{document}
	\onecolumn
	\firstpage{1}
	\title[Hybrid leg compliance]{Hybrid leg compliance enables robots to operate with sensorimotor delays and low control update frequencies} 
	\author[\firstAuthorLast ]{\Authors} %This field will be automatically populated
	\address{} %This field will be automatically populated
	\correspondance{} %This field will be automatically populated
	\extraAuth{}% If there are more than 1 corresponding author, comment this line and uncomment the next one.
	%\extraAuth{corresponding Author2 \\ Laboratory X2, Institute X2, Department X2, Organization X2, Street X2, City X2 , State XX2 (only USA, Canada and Australia), Zip Code2, X2 Country X2, email2@uni2.edu}
	\maketitle
\begin{abstract}
Animals locomote robustly and agile, albeit significant sensorimotor delays of their nervous system. The sensorimotor control of legged robots is implemented with much higher frequencies— often in the kilohertz range—and sensor and actuator delays in the low millisecond range. But especially at harsh impacts with unknown touch-down timing, legged robots show unstable controller behaviors, while animals are seemingly not impacted. Here we examine this discrepancy and suggest a hybrid robotic leg and controller design. We implemented a physical, parallel joint compliance dimensioned in combination with an active, virtual leg length controller. We present an extensive set of systematic experiments both in computer simulation and hardware. Our hybrid leg and controller design shows previously unseen robustness, in the presence of sensorimotor delays up to 60 ms, or control frequencies as low as 20 Hz, for a drop landing task from 1.3 leg lengths high and with a passive compliance ratio of 0.7. In computer simulations, we report successful drop-landings of the hybrid compliant leg from 3.8 leg lengths (1.2 m) for a 2 kg quadruped robot with 100 Hz control frequency and a sensorimotor delay of 35 ms. The results of our presented hybrid leg design and control provide a further explanation for the performance robustness of animals, and the resulting discrepancy between animals and legged robots. 
\keyFont{ \section{Keywords:} legged robots, parallel and passive compliance, hybrid actuation and leg design, sensorimotor delay, feedback, latency}
 %All article types: you may provide up to 8 keywords; at least 5 are mandatory.
\end{abstract}
	
\section{Introduction}
Animals make use of their complex networks of mono-articulate and multi-articulate muscle-tendons to create joint torque and work for bodyweight support~\citep{biewener_scaling_1989}. The time until the animal's actuator (muscle) responds to an external stimulus depends on the velocity and other parameters of nerve conductivity. The delay is 30 ms and higher in a cat-sized animal~\citep{more2018scaling}. In comparison, the entire stance phase at 4 Hz running with a duty factor of 0.4 lasts only 100 ms. Yet animals are seemingly unaffected by the significant delay governing their neuromuscular control, especially at the arguably harshest locomotion event; the touch-down impact. Instead, evidence shows that running birds traverse unforeseen pot-hole perturbations with ease~\citep{daley_running_2006}.

On the other end is an emerging system of latest-generation legged robots driven by `proprioceptive' actuation and control~\citep{park_2017}, `quasi-direct-drive' strategies~\citep{ding_design_2017}, and alike. These robots are capable of real dynamic behaviors and agile maneuvers, with high jumps and landings and fast locomotion~\citep{grimminger2020open,park_2017}. These systems require high-frequency control loops of 500 Hz and more with communication and control delays of a few milliseconds~\citep{bledt_contact_2018,grimminger2020open,li2020centroidal}. Such fully actuated robots consume energy even for a simple, standing task. Proprioceptive and other legged robot designs work well, especially during the stance phase, but they show limitations in the transition from swing to stance phase. Touch-downs can be harsh and include rapid leg loading from zero to multiple body weights in a few milliseconds~\citep{mo_effective_2020}, and side-effects from wobbling masses~\cite{gunther_dealing_2003}. Uncertainties in touch-down timing~\citep{daley2006running} are problematic because the robot's actuators are often gain and impedance scheduled~\citep{hammoud_impedance_2020,hubicki_atrias_2016}. Learned control strategies can mitigate some of the effects of sensor noise and timing uncertainties and also function in unstructured terrain~\citep{bledt_contact_2018}. To function well, the robot's legs must rapidly establish secure and steady ground contact and immediately produce the joint work to support the robot's weight.

In teleoperation, as one of the legged robot applications, time-delays are inherent, and are caused by the communication distance~\citep{varkonyi2014survey}. Sensorimotor delays in legged robotics can be defined more broadly for cases where feedback is transmitted late compared to the expected time-frame to react, i.e., at step-down or push-like perturbations~\citep{daley2018understanding,daley2006running}. Recent hierarchical, optimization-based controllers behave robustly and are versatile, but they heavily depend on the sensorimotor control's exact timing~\citep{shafiee2019online,shafiee2017robust}. In general, force feedback works well in minimal-delay systems, and a large enough feedback delay eventually causes control instabilities. Thus, the communication delay is a significant challenge for legged robots for real-world teleoperation applications. 

Which leads us back to our initial question: how do animal successfully manage an in-built neuromuscular control with large sensorimotor delays, yet show no obvious signs of decline in robustness, responsiveness, or agility.~\citeauthor{more2018scaling} report an exponential relation between the total neuromuscular delay and the animal's mass:
\begin{equation}
    t_{Delay}=0.031 M^{0.21}
\end{equation}
with $M$ as the animal's mass in kilogram, and $t_{Delay}$ as the sensorimotor delay in seconds. For example, for animals with 600 g and 2 kg body weight, the sensorimotor delays are 27 ms and 35 ms, respectively.

Inspired by the animal locomotion apparatus, legged robots increasingly apply designs with passive and active compliance \citep{ambrose2020improved,nasiri2016adaptation}. Designs featuring `series-elastic actuation' (\citep{pratt1995series}) can improve energy efficiency, robustness, and interaction safety\citep{calanca2015review,hutter2011scarleth} and protect the robot's actuators. 
Compliant and elastic parts applied in-parallel to the actuator system are labeled `parallel-elastic actuation' \citep{ambrose2020improved,plooij2016reducing,niehues2015compliance,roozing2018modeling,ruppert2019series,sprowitz2013towards}. 

Legged robots with in-series and parallel joint elasticity can locomote purely driven by feed-forward control, without sensor feedback informing the controller of the robot's touch-down status, or the overall robot state \citep{sprowitz2013towards,narioka_development_2012,ruppert2019series,sprowitz_oncilla_2018}. Compliance in-parallel to the actuator system can improve energy efficiency~\citep{roozing2019efficient,liu2018switchable,yesilevskiy2018energy}, and increase the system's stability~\citep{ahmadsharbafi2020parallel}. Most important for the touch-down event, parallel compliance provides an \emph{immediate, physical} response from its spring-equipped leg. The leg's spring will charge without delay, sensory feedback or control input, and it will carry the robot's weight. Legged robots with in-parallel leg designs have been shown to mitigate step-down perturbations similarly to running animals~\citep{sprowitz2013towards,daley2006running}.
But compliant elements cause under-actuation and therefore lower the robot's active control authority.  In parallel-elastic systems, some amount of control authority in task-space remains, but the robot's actuators will work against its full-strength, in-parallel elasticities.

Animals are different in many ways, and it seems they only benefit from combined passive and active compliant leg actuation~\citep{alexander1990three}. Physical, compliant structures allow a design where the control task is partially taken over by the body's mechanics~\citep{blickhan2007intelligence}. Keeping the animals' discrepancy between neuromuscular control delays~\citep{more2018scaling,more2010scaling} and high locomotion robustness in mind, we wondered if these aspects are related.

We developed a hybrid between two systems, and merged what we see in legged animals, to gain the best of two systems; 
A) Passively, spring-loaded legged robots that work well with open-loop control, i.e., without control feedback, and
B) High-bandwidth, fully actuated legged robots with full control authority and rapid response times.

Inspired by biological systems, we show a hybrid robot mechanism and control design with \emph{complementing levels of passive and active joint compliance}. The concept successfully overcomes significant sensorimotor delays and works with lower sensorimotor control update frequencies, than the state-of-the-art fully actuated legged robots working with delays in the low millisecond range \citep{grimminger2020open,park_2017}. Our hybrid robot leg draws comparatively lower actuator power due to its partial, in-parallel passive compliance. The proposed hybrid design leads to a balanced level of control authority, in-between that of robots with passively compliant legs and full state controlled robots.

In \Cref{sec:two}, we present a theoretical stability analysis of a simplified joint design with hybrid passive and active stiffness in the presence of sensorimotor delays. We then present extensive computer simulation and hardware experiments and investigate the effect of varying actuation update control frequencies and sensorimotor delays on a robotic leg with varying ratios of passive and active stiffness and a simulated quadruped robot equipped with these legs (\Cref{sec:three}). We conclude our work in \Cref{sec:four}.

\section{Materials and methods}
\label{sec:two}
\begin{figure}[h!]
\centering
\includegraphics[width=.8\linewidth]{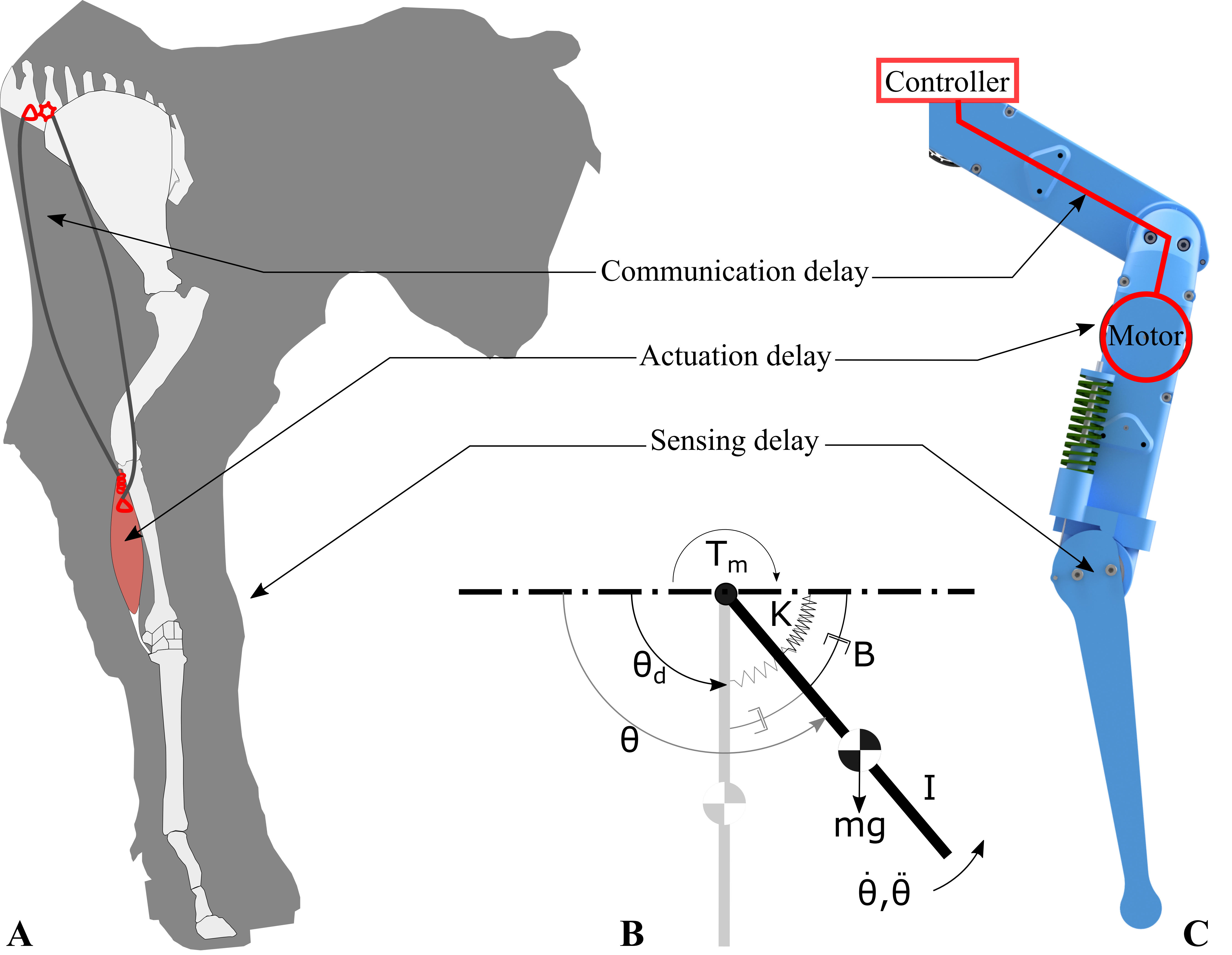}
\caption{Different source of delay in the robotic leg and biological counterpart and modeling the leg as a simple pendulum.}
\label{fig:goatSchematic}
\end{figure}
We first analyse the poles of a linear and simplified actuated system with sensorimotor delays. Its mechanics consist of a pendulum mounted with parallel compliant elements~(\Cref{fig:goatSchematic}B). We then implement hardware experiments and simulations to investigate the effect of hybrid active and passive compliance on a multi-body leg's control performance. We quantify the total (sum of) system compliance as the active compliance in-parallel to the passive (spring-based) compliance, acting at a robotic knee joint (\Cref{fig:goatSchematic}C):
\begin{equation}
K_{Total} = K_{Active} +K_{Passive}
\end{equation}
Where $K_{Passive}$($ \frac{Nm}{rad}$) is the joint's passive rotational stiffness, $K_{Active}$($ \frac{Nm}{rad}$) is the joint's active (virtual) rotational stiffness ($ \frac{Nm}{rad}$) provided by its actuator. $K_{Total}$($ \frac{Nm}{rad}$) is the total rotational stiffness of the joint. We define the ratio $\lambda_{Passive}$ as the passive stiffness over the total stiffness. 
\begin{equation}
\lambda_{Passive}  = \frac{K_{Passive}}{K_{Total}}
\label{eqn:ratio}
\end{equation}
For example, a low passive compliant ratio of \num{0.1} indicates that the knee spring supplies \SI{10}{\%} of the knee's stiffness to carry the robot, and the active, virtual leg compliance supplies the remaining \SI{90}{\%}.

\subsection{Theoretical analysis of a reduced order model}
We analysed a simplified system without impacts, with a load and parallel compliance, to analytically quantify the effects of sensorimotor delays. The reduced order model consists of a strut-leg mounted as a single degree-of-freedom pendulum (\Cref{fig:goatSchematic}B), and represents a robot's lower leg. The equations governing the pendulum motion are:
\begin{equation}
\label{eq:pendulum_dynamics1}
I \ddot \theta + m g l \cdot sin (\theta - \theta_{d}) +K_{Passive} (\theta - \theta_{d})+B (\dot \theta - \dot \theta_{d})= \tau_{knee}
\end{equation}
where $B$ is the physical system's damping, $K_{Passive}$ is the stiffness of the parallel compliance element, $I$ is the momentum of inertia, $m$ is the mass, $l$ is the Center of Mass distance to the pivot point, $g$ is the standard acceleration. $\theta_d$ is the equilibrium knee joint angle that corresponds to the orientation that a relaxed spring, $\theta$ is a joint angle and $\tau_{knee}$ is the control torque input to the knee joint. We implement the input torque as an active compliance:
\begin{equation}
\label{eq:pendulum_dynamics2}
I \ddot \theta + m g l \cdot sin (\theta - \theta_{d}) +K_{Passive} (\theta - \theta_{d}) +B (\dot \theta - \dot \theta_{d})= - K_{Active} (\theta_{feedback} - \theta_{d})
\end{equation}
where the $K_{Active}$ is related to the active compliance from motor torque. $\theta_{feedback}$ is the joint angle read by sensor. We assume a small enough angular deviation of the pendulum around the equilibrium point: $ sin (\theta - \theta_{d}) \simeq (\theta - \theta_{d}) $, and can write \Cref{eq:pendulum_dynamics2} as a linear differential equation. We convert \Cref{eq:pendulum_dynamics2} to the Laplace domain and incorporate a fixed time delay $t_d$ in the feedback loop of the control input (active compliance). The resulting closed loop system transfer function can be presented in the frequency domain as:
\begin{equation}
\label{eq:pendulum_laplace}
\frac{ \Theta_s}{\Theta_{ds}}=\frac{ K_{Active}  e^{-t_d s}+m g l +K_{Passive}}{s^2 I + B s + K_{Active}  e^{-t_d s} +K_{Passive}+ m g l}
\end{equation}
The displacement of the system's poles can be analyzed to understand the effects that a combination of active and passive compliance have, on the closed-loop stability in the presence of sensorimotor delay. We linearised the system's exponential time delay term with a third-order Pad\'e approximation. The desired, full joint stiffness $K_{Total}$ is achieved with the combination of active and passive joint compliance (\Cref{eqn:ratio}).

\begin{figure}[h!]
\centering
\includegraphics[width=.95\textwidth]{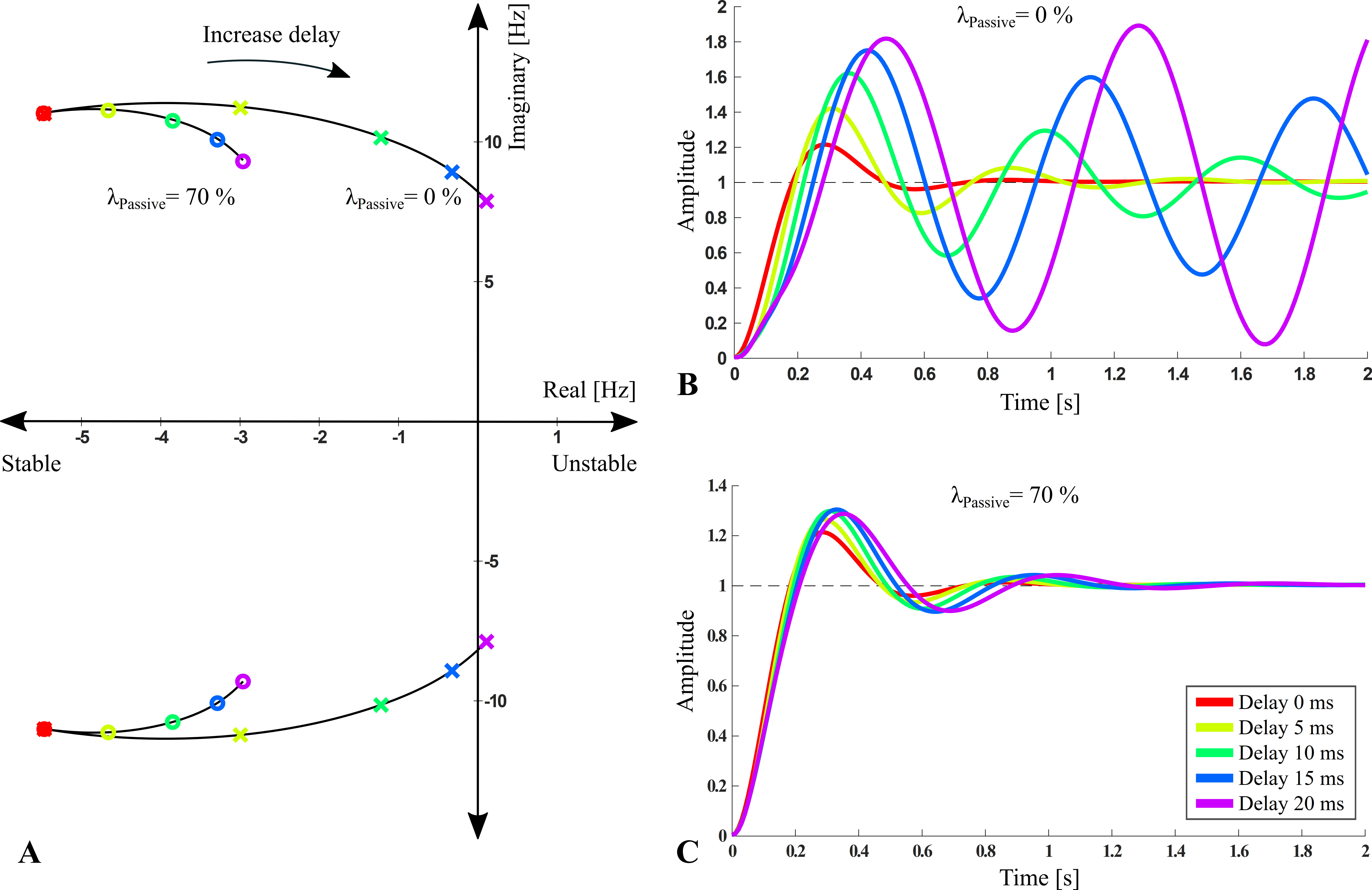}
\caption{%
A) Pole analysis of a simplified pendulum system with in-parallel, passive and active joint elasticity. The effects of varying amounts of delay and passive compliance on the stability of the system are shown. Here, the system's mass is $m=0.5\;kg$, the total joint stiffness is $K_{total}=1.15\;\frac{Nm}{rad}$, and the damping coefficient is $B = 0.14\;\frac{Nms}{rad}$. 
B) The system's step response for varying delays with $\lambda_{Passive}= 0$. 
C) The system's step response for varying delays with $\lambda_{Passive}= 0.7$.}
\label{fig:stabiltyAnalysis}
\end{figure}

Our analysis of the system's poles is illustrated in \Cref{fig:stabiltyAnalysis}A. The $"\times"$ mark shows cases where the joint's compliance is fully active, i.e., $\lambda_{Passive}=0 $. The $"\circ"$ mark indicates cases where $70 \%$ of the total compliance is caused by the spring, i.e., $\lambda_{Passive} =0.7$. 
\Cref{fig:stabiltyAnalysis}A shows that in a case of $\lambda_{Passive} =0$, and with increasing feedback loop delay, the dominant system poles move rapidly from their stable region towards the unstable region at the imaginary axis. For passive compliance in-parallel with active actuator compliance, the rate of divergence is much lower. The system's step response (\Cref{fig:stabiltyAnalysis}B) indicates that increasing the sensorimotor delay with a $\lambda_{Passive} =0 $  leads to high oscillations, and a resonance effects eventually destabilizes the joint. However, in the case of combined passive and active stiffness with an increased delay of $20 ms$, the closed-loop system's step response  is stable and smooth (\Cref{fig:stabiltyAnalysis}C, purple line). 

\hspace{0.5cm}
\subsection{Computer simulating articulated robot legs}
The previous results from the analysis of poles of a simplified, linearised system  provided insights about sensorimotor systems under considerable sensorimotor delay, and hybrid joint compliance. Here we are extending our characterization to an articulated leg with hybrid joint compliance, sensorimotor delays, and impacts from touch-down. The landing task is one of the most challenging motions due to the large impulsive ground reaction force applied at the system. The landing load case is nonlinear, hybrid, under-actuated, and basically a system's step response. A step response analysis like the drop-experiment applied here is one conventional way for a system characterization, in control theory. We implemented our simulation in the Pybullet simulator~\citep{coumans2019}, and we performed extensive computer simulations for the landing task, for a broad range of sensorimotor delays, frequencies, and varying $\lambda_{Passive}$ values. We simulated a single leg and a quadruped robot, both modified from the open-source quadruped robot 'Solo'~\citep{grimminger2020open}. 

\begin{figure}[h!]
\centering
\includegraphics[width=.9\textwidth]{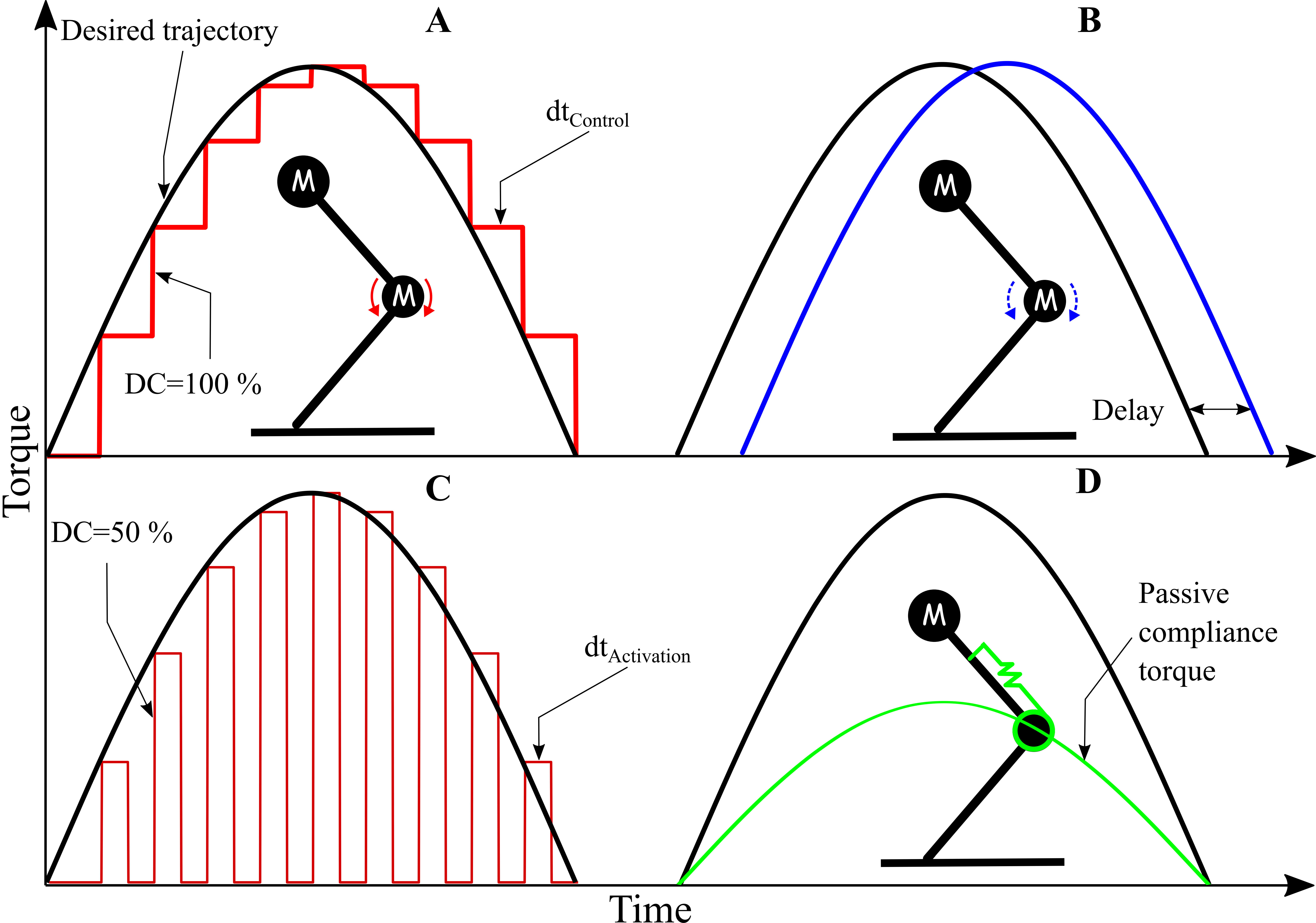}
\caption{Knee motor command for different combinations of control frequency, frequency duty cycle, and sensorimotor delay. 
A) Is showing a \SI{100}{\%} duty cycle control frequency, at a relatively low update frequency.
B) Shows a set sensorimotor delay between the desired knee output torque, and the commanded output torque.
C) Shows a \SI{50}{\%} duty cycle at the same control frequency as A). 
D) Indicates the portion of parallel, passive joint compliance, of the joint's overall joint compliance. Here, a passive compliance ratio of $\lambda_{Passive}$ 0.5 is shown, such that the knee's mechanical spring produces half of the knee torque, and the virtual knee spring produces the remaining knee torque.
}
\label{fig:motorcommand}
\end{figure}

In \Cref{fig:motorcommand}, we depict the tested control and sensorimotor strategies. The black curve is a schematic, desired knee motor torque trajectory, and the control command frequency (red line) is measured in commands per second. For a reference; the control frequency of active actuators in proprioceptive legged robot is often around 1 kHz, i.e., a cycle period  $dt_{control}= \frac{1}{freq}$ of 1 ms. Here we are especially interested in investigating scenarios with control frequencies much below 1 kHz. We defined three strategies for applying torque with a duration of $dt_{Activation}$, and a duty cycle $DC$, which is the fraction of $dt_{control}$ with a non-zero actuator torque. $dt_{Activation}$ is defined as the time period between control commands, i.e., $dt_{Activation}= DC \times dt_{control}$, and ranges from \SI{1}{ms} to a maximum of $\frac{1}{freq}$ in \si{[ms]}. For $dt_{Activation,min}$, the control command is applied for \SI{1}{ms}, and then reset to zero. For $dt_{Activation,max}$, the actuator will send a command equal to the previous value until the control command is updated (\Cref{fig:motorcommand}, red line). We further simulate cases where the control command is applied after a given sensorimotor \textit{delay} (\Cref{fig:motorcommand}, blue curve).  

The active compliance controller input is applied at the knee joint as:
\begin{equation}
\label{eq:motorTorque}
\tau_{knee,motor}= K_{Total} (1-\lambda_{Passive})   (\theta_{feedback, Knee} - \theta_{d, knee}) 
\end{equation}
To present the physical spring in Pybullet, we implement a spring torque applied at the knee joint:
\begin{equation}
\label{eq:SpringTorque}
\tau_{knee,spring}= K_{Total} (\lambda_{Passive})   (\theta_{ knee} - \theta_{d, knee})  
\end{equation}

\subsection{Setup hardware experiments}
\begin{figure}[h!]
	\centering
	\includegraphics[width=.9\linewidth]{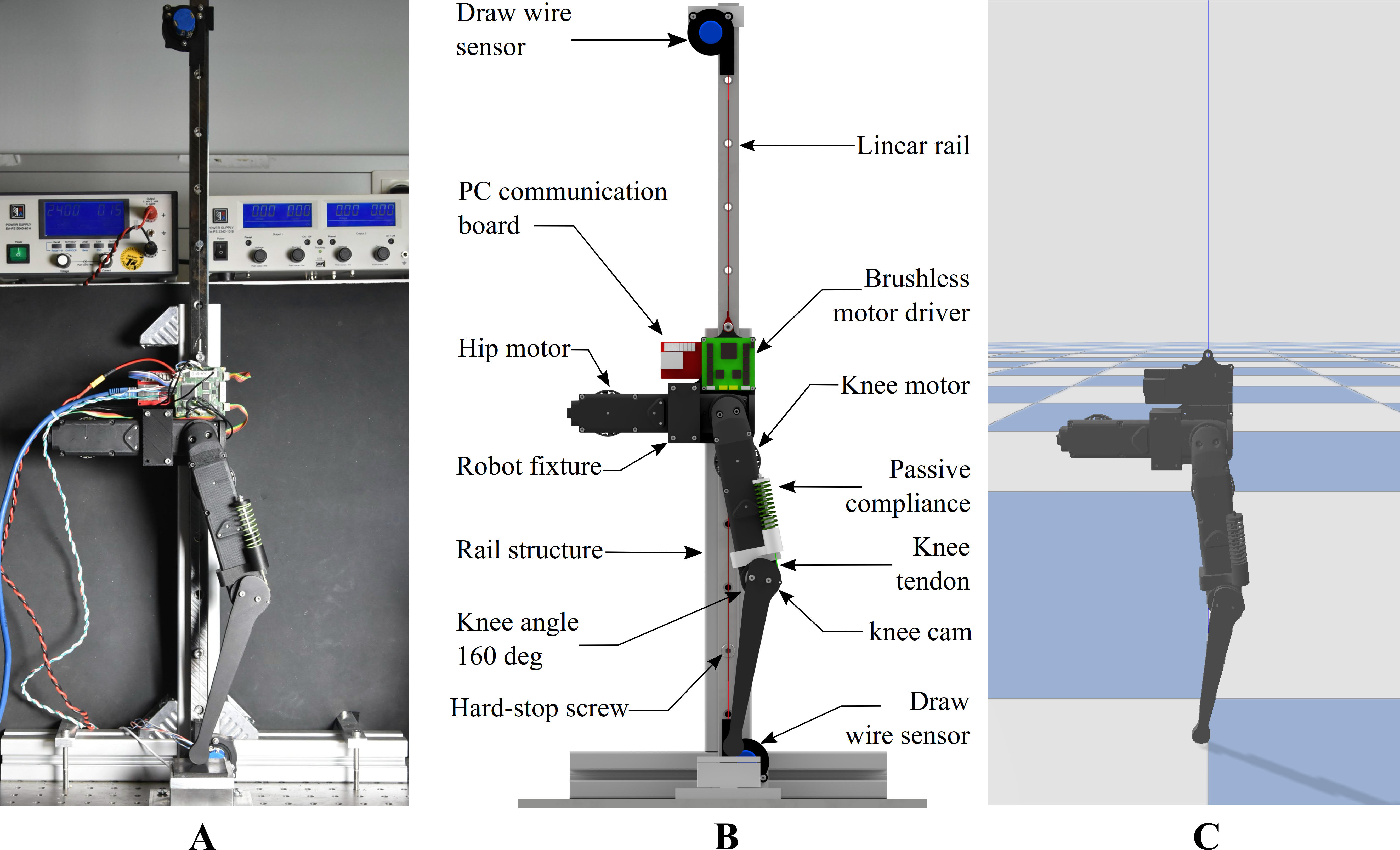}
	\caption{Experimental setup. 
A) A 2-DoF hybrid compliant leg; a one-directional spring (`passive compliance') extends the knee joint via a knee tendon and a knee pulley/cam. Knee springs with varying stiffness were mounted during the experiments, supporting between \SIrange{0}{100}{\%} of the robot's weight. A rail guides the robot's vertical drop, and a set of potentiometers measures the robot's height. The leg's active knee actuation acts in combination to the in-parallel mounted knee spring.
B) Setup details. 
C) The URDF model of the hybrid compliant robot leg in the Pybullet simulation.}
	\label{fig:experimentFigure}
\end{figure}

We modified a single leg of the 8-DoF open-source, quadruped robot 'Solo' with its proprioceptive design and controller architecture~\citep{grimminger2020open}.
The leg has two active degrees of freedom, one at the hip and one at the knee. Both leg segments are $0.16 \: m$ long.  A brushless motor (Antigravity MN4004-kv380, \textit{T-Motor}) drives a two-stage belt transmission with an overall $9:1$  gear ratio. Two encoders (AEDT-9810-T00, \textit{Avago}) measure the motor's rotor position, which is recalculated into joint angles. We added a physical spring (SWY 16.5-30, \textit{Misumi}) in parallel to the knee. The spring is inserted into the knee joint via a tendon and a pulley with \SI{18.9}{mm} radius (\Cref{fig:goatSchematic}C). We designed the spring mount to rapidly exchange springs by softer or harder ones, between experiments. To simplify the touch-down scenario, the robot leg was dropped while guided by a vertical rail (\Cref{fig:experimentFigure}A). The hip joint is constrained to follow half of the knee joint angle at all times, controlled by a position controller, and is meant to create foot contact vertically below the hip joint. We monitor the robot's vertical position with a  draw-wire sensor (LX-PA-40, \textit{WayCon}), we use two draw-wire sensors on top and bottom to cancel unwanted tension force produce by draw-wire sensors. With the vertical robot leg position, we quantify the robot's landing behavior and evaluate the effectiveness of the proposed hybrid active/passive joint stiffness framework. The recorded vertical position is sampled by an analog-to-digital (A/D) port of the brushless motor driver board. The brushless motor driver board sends motor position and vertical position data via a communication board to a PC, and it receives the motor control commands, via a SPI Protocol. The PC communication board is the bridge between the brushless motor driver board and the PC (Intel® Xeon(R) W-2145 CPU @ 3.70GHz $\times$ 16, 64-bit, RAM 62,5 GB, Ubuntu 18.04) via EtherCAT. We wrote a custom robot control program in a Python wrapper, which communicates via the PC communication board with the robot leg. The Python wrapper adds a time stamp to all input parameters and sensory data such as joint angles, motor currents, and body height, and saves all data into a text file for further analysis. A custom Matlab script was written to post-process and plot the data.

\section{Result and discussion}\label{sec:three}
In this section we present the results from the computer simulation of a single robot leg with hybrid joint compliance, and from dropping an equivalent quadruped robot in simulation. We then present the results from our hardware experiments with a single leg mounted to a vertical slider.

\subsection{Single leg computer simulation}
We study the effects of varying combinations of sensorimotor delay, control loop update frequency, and the ratio of passive compliance $\lambda_{Passive}$ on the compliance controller performance during landing. 
We derive a reference landing hip height trajectory from dropping a parallel compliant leg with $\lambda_{Passive}=1$. Deviations in settling time and final hip height are accepted as successful landings, within given limits.

We performed computer simulations to quantify the viability of the landing task. We varied $\lambda_{Passive}$ from $0$ to $1$ by steps of $0.05$, the sensorimotor delay from $0 \: ms$ to $60 \: ms$ in steps  $5 \: ms$, and the sensorimotor control loop update frequency in a range of $[ 20,\: 50,\: 250, \:1000]\:Hz $.  
In Pybullet, we set joint damping values of $0.01 \frac{Nms}{rad}$ and  $0.05 \frac{Nms}{rad}$ for the hip and knee, respectively. The weight of the single robotic leg is $0.6 \: kg$, the weight of the quadruped is $2.0 \: kg$. We chose the total stiffness for the knee joint compliance for an approximate leg length deflection of $10 \%$ during mid-stance, after dropping the leg from a height of $42.5 \: cm$. We also assumed that the robot's weight is concentrated at its hip joint.
We realized a $\lambda_{Passive}=1$ rotational spring as a combination of a pulley with radius $ r=0.0189 \: m$, and a linear spring with a stiffness of $ K=4680 \: \frac{N}{m}$ acting at the knee joint. The total rotational stiffness is then $K r^2=1.67 \: \frac{Nm}{rad}$ as a combination of passive and active compliance components. In simulation, the delay and frequency modes do not affect the $\tau_{knee,spring}$, as it is representing a physical spring. Instead, delay and frequency variations only affected the active part of the joint stiffness $\tau_{knee,motor}$. We dropped the robot from $0.425 \:m$ height, which is 1.3 times the length of both leg segments. The robot's body was constrained to move vertically only, and we monitored the robot's height during landing. The vertical hip trajectory represents the system's step response. We then assessed the robot's controller performance by measuring the settling time and the final hip height. The settling time is the duration to settle to the hip height within $| \mathrm{Height(t) - Height_{final}}|$ range, between the hip height trajectory $\mathrm{Height(t)}$ and the steady-state hip height $\mathrm{Height_{final}}$. 
The reference final hip height is that of a $\lambda_{Passive}=1$ robot leg.
We chose a critical settling time above $0.7 \: s$, and a critical hip height below $0.3 \: m$ as failing cases. 

\begin{figure}[h!]
\centering
\includegraphics[width=1.0\textwidth]{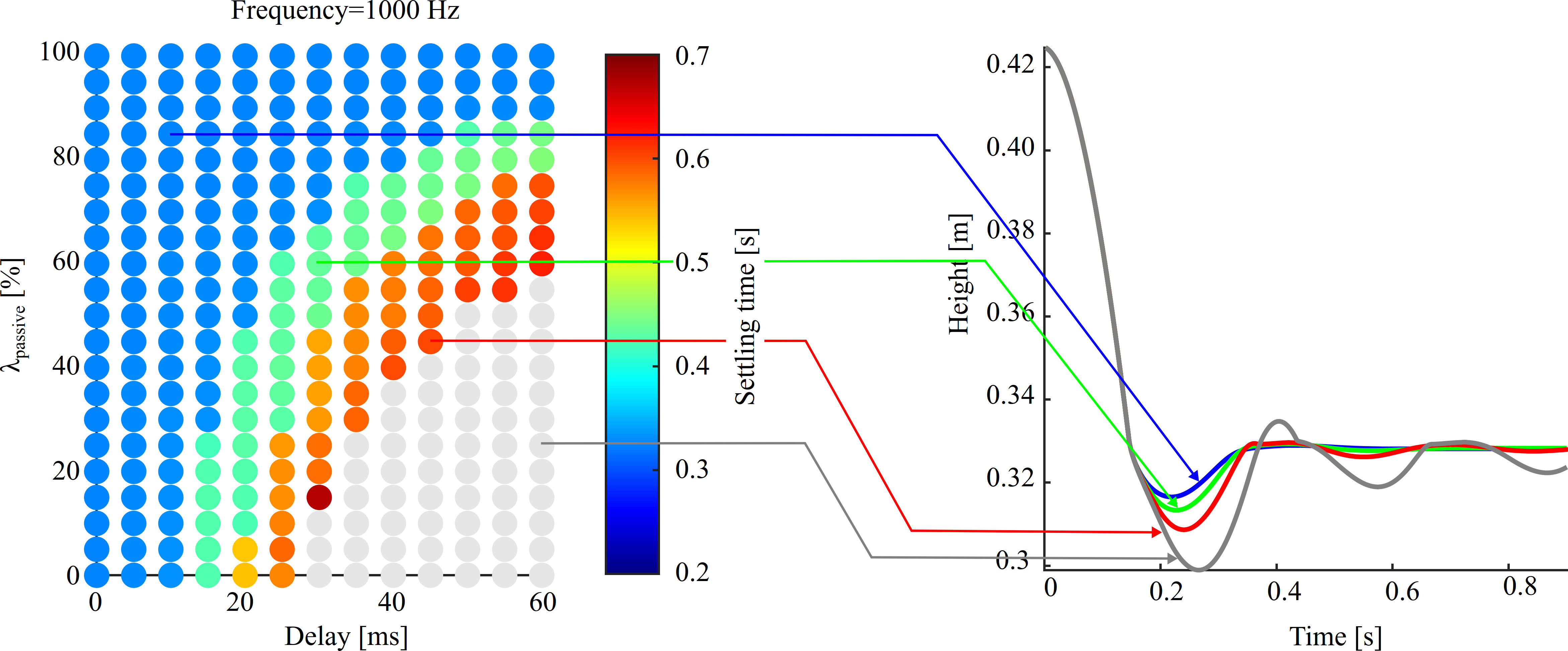}
\caption{Results from 315 computer simulations for a control update frequency of $1000\;Hz$ for the drop-landing task of a simulated robot leg. 
The passive compliant ratio $\lambda_{passive}$ was varied between $0$ to $1$ in steps of $0.05$, and the sensorimotor delay between $0\:ms$ to $60\:ms$ by steps $5\:ms$. The grey data points and the grey hip height trajectory show a failed landing task with too much initial deflection and a too large settling time. The colored data points and trajectories show viable landings. Viable landings are shown for sensorimotor delays up to \SI{60}{ms}, in combination with a passive compliant ratio of $\lambda_{Passive}=0.6$.
}
\label{fig:result1000Hz}
\end{figure}

In \Cref{fig:result1000Hz}, the results of 315 drop-landing simulations are illustrated, with varying sensorimotor delays and $\lambda_{Passive}$ settings, and for a single control frequency of $1 \: kHz$.
The grey points represent failed simulation cases with a steady-state hip height of less than $0.3 \: m$ or settling times higher than $0.7 \: s$. Our computer simulation results show that in a case of $\lambda_{Passive}=0.0$, and by increasing the sensorimotor delay to values above $25\: ms$, all scenarios failed. Yet by setting $\lambda_{Passive}$ high than $0.7$, the system is stable at the presence of $60 \: ms$ delays. The computer simulation results demonstrate that the hybrid compliant leg has stable impact control regimes in the presence of large sensorimotor delays, with an appropriate combination of passive and active compliance.

\begin{figure}[h!]
\centering
\includegraphics[width=1.0\textwidth]{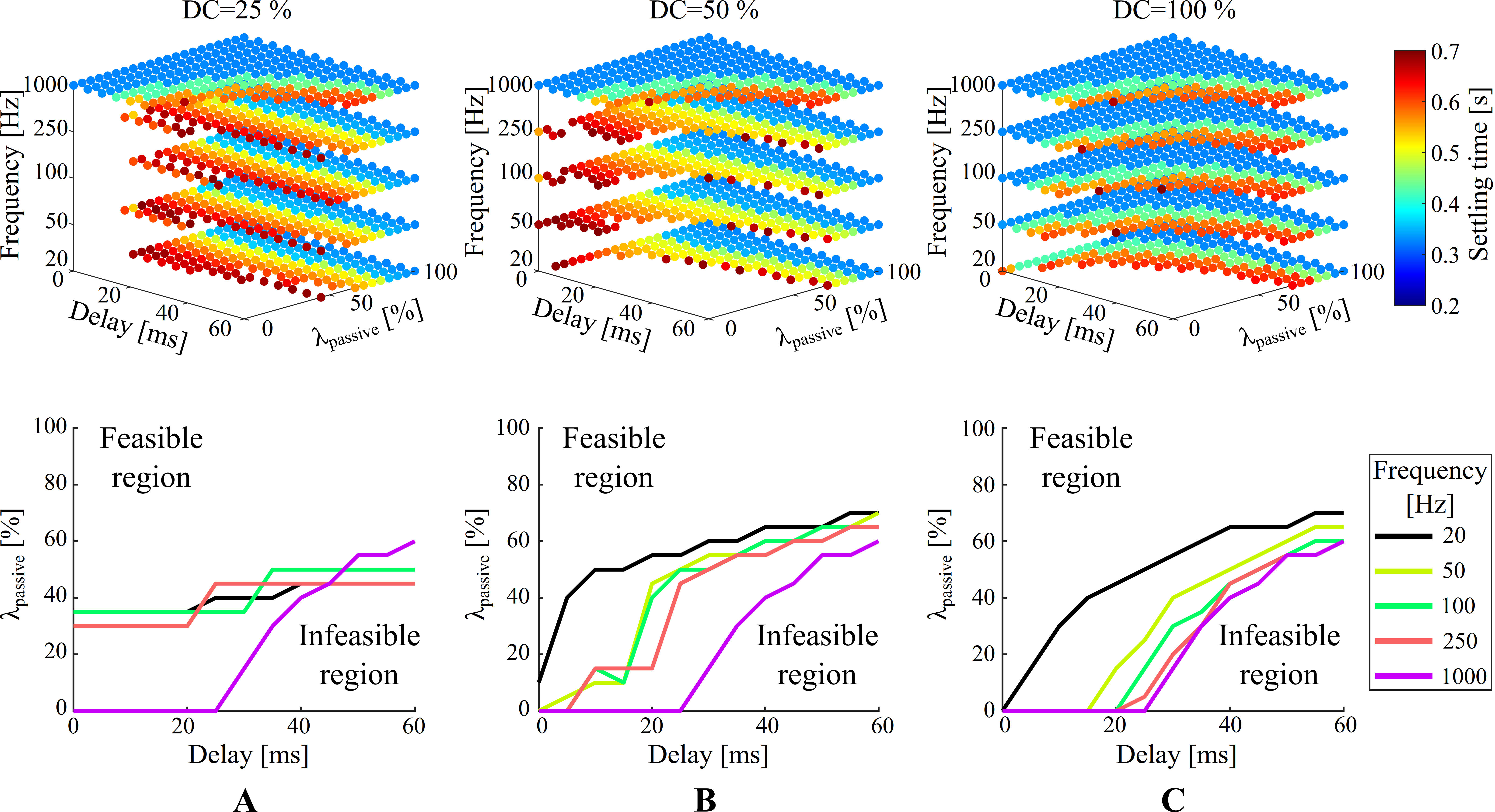}
\caption{Results showing three different duty cycles (DC). Each DC plot is made from 1575 simulations, in sum 4725 computer simulations. Top and bottom plots show the same data. The reference landing performance is the top left data point in each plot, from dropping a $\lambda_{Passive}=1$ robot leg.
A) $DC =25 \: \%$  
B) $DC =50 \: \%$  
C) $DC =100 \: \%$.
}
\label{fig:result}
\end{figure}

We then investigated the effect of varying the control update frequency, and we performed computer simulations for frequencies $[ 20,\: 50, 100\:, \: 250, \:1000]\:Hz $, and duty cycles of $[25,\;50,\;100]\:\% $. The results are shown in \Cref{fig:result}.  
Most visible is a decreasing viable area for all three duty cycles, from reduced the control frequencies.
Comparing duty cycles of $25\%$, $100\;\%$ (\Cref{fig:result}A, \Cref{fig:result}C) shows that for a reduced duty cycle, the viable areas did overall change much. Low amounts of $\lambda_{Passive} \approx 0.2$ only lead to viable landings in combination with a DC of \SI{50}{\%} or the highest update frequency (\SI{1}{kHz}).
The \Cref{fig:result}C shows that a DC of $100\: \%$ at almost all control frequencies $[\: 100,\: 250, \:1000]\:Hz $ have a similar viable area. Only when switching to control frequencies as low as $20\: Hz$, the viable area largely changes.  
For a DC of $50\;\%$ (\Cref{fig:result}B), the viable area changes slightly when switching between control frequencies $[\: 50,\: 100,\: 250]\:Hz $.
The biggest viable area changes are visible when changing from $1000\:Hz$ to $250\:Hz$, or from $50\:Hz$ to $20$.
At a DC of $25\;\%$ (\Cref{fig:result}A) and when switching between control frequencies of $1000\: Hz$ to $250\: Hz$, the viable area is reduced drastically. 
The change of control frequency for the remaining values has seemingly less influence on the change of viable area. 
For control frequencies of $20, \:50,\:100,\:250 \: Hz $ and high sensorimotor delays around $50, \: 60 \:ms$, and by reducing the duty cycle, a larger viable area for large $\lambda_{passive}$ is visible.
For all $\lambda_{Passive}$ above $0.6$ we show stable landing, including critical combinations of $60 \: ms$ delay and $20 \: Hz $ updated frequency. These examples show how capable a hybrid active and passive compliance system can be. In general, all results indicate stable landing for passive compliant ratios equal and higher than $\lambda_{Passive}$ $0.7$.

\subsection{Quadruped computer simulation}
\begin{figure}[h!]
\centering
\includegraphics[scale=0.53, trim ={0.1cm 0.1cm 0.01cm 0.01cm},clip]{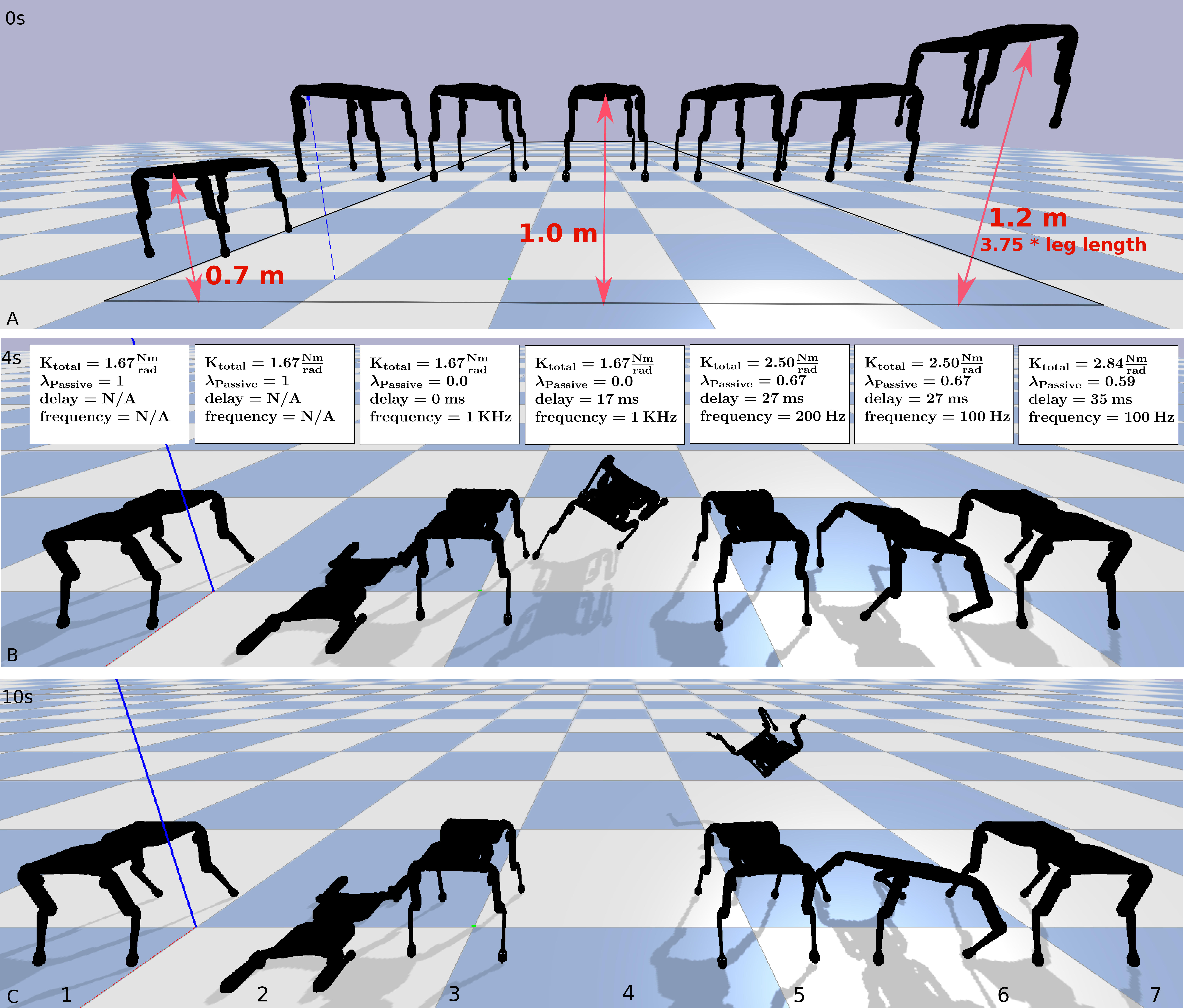}
\caption{Computer simulated quadruped robots landing, in seven different scenarios. 
A) The Robot's initial state. The initial drop heights are indicated in red. 
B) An intermediate robot state at 4 s simulation time. The panels also provide controller parameters.
C) Converged robot state after 10 s. Simulation cases 1, 3, 5, and 7 were landing successfully.}
\label{fig:simulationQuadruped}
\end{figure}

The previous computer simulation results from landing a single leg indicate that by choosing a high ratio of passive compliance, the robot's performance becomes largely independent of the sensorimotor delay, and the control frequency.
But fully passive compliance limits a robot's agility, and to maintain control authority we suggest reducing the passive compliance ratio.
In seven drop-landing scenarios, we altered a quadruped robot's drop height, and its hybrid passive and active stiffness (\Cref{fig:simulationQuadruped}). The simulation parameters are provided in \Cref{tab:case}.
The robot in case 1 features a passive compliance ratio of \num{1}. It was dropped from a height of $0.7 \: m$, and landed successfully.
The robot in case 2 uses identical parameters, and it was dropped from $1 \: m$ height, and fails to land properly. This example shows the drawback of pure, passive and non-adjustable compliance.
The robot configuration in case 3 is a controller with full, bi-directionally active compliance, no passive compliance, and no sensorimotor motor delay. The robot's controller successfully guided the landing.
In case 4 a fully-active compliance with a $17 \: ms$ sensorimotor delay failed to land properly, which shows the vulnerability of active compliance in the presence of sensorimotor delay. 
Case 5 shows a successful landing scenario by combining passive and active compliance, with a $27 \: ms$ sensorimotor delay. The control update frequency was reduced from $1000 \: Hz$ to $200 \: Hz$. 
Case 6 uses the same passive compliance ratio $\lambda_{Passive}$ as case 5, and we reduced the control update frequency to $100 \: Hz$. The robot's controller did not land the robot successfully.
Finally in case 7, we decreased the passive compliance ratio $\lambda_{Passive}$ by increasing the active stiffness, and the robot landed successfully, for a drop height of $1.2 \: m$, a sensorimotor delay of $35 \:ms $, and $100 \:Hz$ control frequency. The last case shows how an appropriate combination of hybrid active and passive compliance at low control frequency increases the control authority, energy efficiency (the spring partially captures the robot's load) and robustness in the presence of sensorimotor delay.

\vspace{0.5cm}
\subsection{Hardware experiments}
\begin{figure}[h!]
\centering
\includegraphics[width=.9\linewidth]{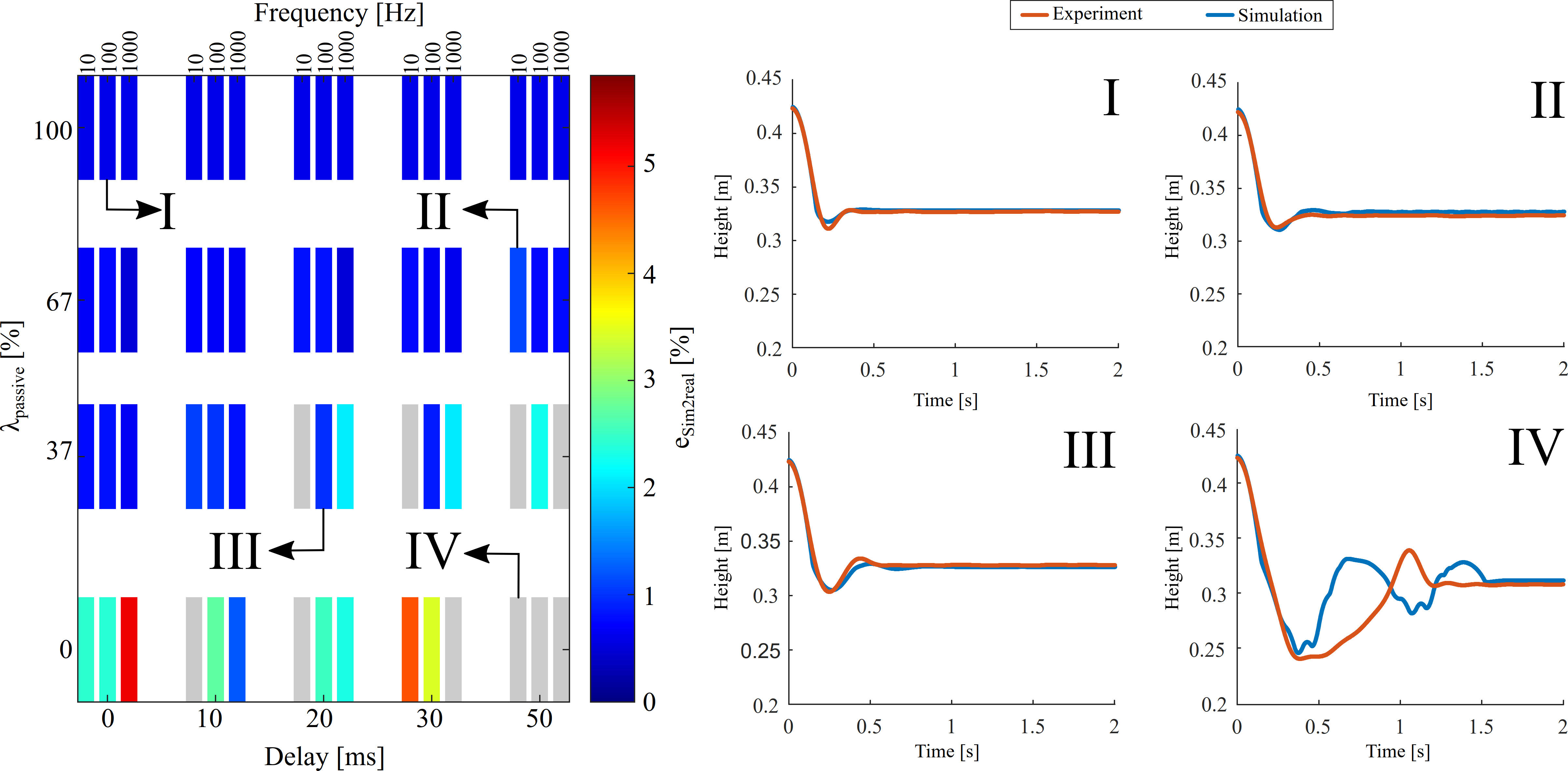}
\caption{Comparing results from computer simulation and hardware experiments, in form of a mean-square-error of the normalized body height. Left) Good similarities are shown as colored data patches, grey data patches indicate unviable landing experiments. I-IV) exemplary hip trajectories for common control frequency and passive compliant ratio combinations. I-III are successful landings with short settling times and a good final hip height. IV shows an unsuccessful landing example.}
\label{fig:errorSim2Real}
\end{figure}

We validated previous simulation results with hardware experiments. We selected passive compliance ratios of $\lambda_{Passive}=[0 \:,0.37\:,0.67\:,1]$ and a total knee stiffness of $K_{Total}=4680  \:Nm$. We then varied the physical spring stiffness, control frequencies  $Frequencies=[1000,\:100,\: 10]$ and sensorimotor delays $\:Delays = [0,\:10,\: 20,\:30,\:50] \si{ms}$. 
In \Cref{fig:errorSim2Real} we assess the difference $e_{sim2real}$ between the computer simulation results and the hardware experimental results, as the mean-square-error between both resulting vertical hip trajectory, normalized by the maximum leg length.
Grey color data shows failure cases in both experiments and simulations. Viable cases with a mean-square-error of less than \SI{6}{\%} (\Cref{fig:errorSim2Real}, deep red) indicate a good consistency between the hardware experiment, and the computer simulation. We show four exemplary hip trajectory, with varying passive compliance ratios $\lambda_{passive}$ (\Cref{fig:errorSim2Real}, I-IV). 
The first three cases are viable cases with a good consistency between simulation and experiments, and with a $e_{sim2real}$ of less than \SI{6}{\%}. The fourth case is a typical failure case. It shows an atypical settling and transition behavior both in the computer simulation and in the hardware experiment. 

\section{Conclusion and summary} \label{sec:four}
We proposed a new, hybrid parallel passive and active compliance architecture for the design and control of robotic legs, where a portion of the leg loading is captured by the in-parallel knee spring. The remaining load is captured by the robot's active, virtual spring. We show robust landing scenarios with varying hybrid compliance ratios in the presence of sensorimotor delays and low control frequencies. The hybrid compliance design remains relatively energy efficient because the leg's in-parallel spring stores and releases significant elastic energy. Yet with an appropriate amount of active compliance, the robot's controller maintains its control authority.
We performed extensive computer simulations and systematically characterized the system's response to varying sensorimotor delays and control frequencies in a drop-landing task. 
The computer simulation results show that drop-landings were successful for sensorimotor delays up to $60 \: ms$, and a control frequency as low as $20 \: Hz$, in combination with a passive compliance ratio of $\lambda_{Passive}=0.7$.
We also simulated drop-landings with a quadruped robot, with varying total leg stiffness and multiple drop heights. In these examples, the hybrid compliant actuation showed good robustness in the presence of sensorimotor delay and low update control frequencies. With a significant amount of active compliance ratio, we still have good control authority; we altered the robot's total compliance and successful drop-landed the robot from multiple heights. 
Our results are well consistent with the locomotion robustness of running animals, which evidently and successfully deal with neuromuscular sensorimotor delays in a similar range.
We verified our computer simulation results with hardware experiments for a selected range of control and design parameters and could show a good agreement between both.

\section*{Author Contributions}
MSA and AAS contributed to the concept, robot design, experimental setup, simulation, and experimentation. AB-S developed the original hybrid active/passive compliant concept. All authors discussed the data, agreed with the presented results, and contributed to the writing.

\section*{Acknowledgments}
The authors thank the International Max Planck Research School for Intelligent Systems (IMPRS-IS) for supporting AAS. We thank Felix Grimminger and Jad Saud for support developing the robot leg, the Robotic ZWE for prototyping support and Julian Viereck for providing the URDF file of the quadruped.
	
\section*{Supplemental Data}
\href{}{The supplementary material} includes a code algorithms snippet driving the simulations, and our video presenting results from computer simulation and hardware experiment.
	
\section*{Data Availability Statement} The datasets recorded and generated for this study are available on request from the corresponding author.
	
\bibliographystyle{frontiersinSCNS_ENG_HUMS}
\bibliography{References}
	%%% Table should be at the end of paper
	
% 	\begin{table}[h!]
% 		\centering
% 		\resizebox{.83\linewidth}{!}{%
% 			\begin{tabular}{cccccc}
% 				\hline
% 				& \begin{tabular}[c]{@{}c@{}}Total (sum)\\ Compliance\\ {[}N/m{]}\end{tabular} & \begin{tabular}[c]{@{}c@{}}Active\\ Compliance\\ {[}N/m{]}\end{tabular} & \begin{tabular}[c]{@{}c@{}}Passive\\ compliance\\ {[}N/m{]}\end{tabular} & \begin{tabular}[c]{@{}c@{}}Control\\ Frequency\\ {[}Hz{]}\end{tabular} & \begin{tabular}[c]{@{}c@{}}Delay\\ \\ {[}ms{]}\end{tabular} \\ \hline
% 				\textbf{Case 1} & 4680 & 0 & 4680 & 1000 & 0 \\
% 				\textbf{Case 2} & 4680 & 0 & 4680 & 1000 & 0 \\
% 				\textbf{Case 3} & 4680 & 4680 & 0 & 1000 & 0 \\
% 				\textbf{Case 4} & 4680 & 4680 & 0 & 1000 & 17 \\
% 				\textbf{Case 5} & 7020 & 2340 & 4680 & 1000 & 27 \\ \hline
% 			\end{tabular}
% 			\label{tab:case1}}
% 		\caption{Parameters of the experimental case study.}
% 	\end{table}
	
\begin{table}[h!]
	\centering
		\resizebox{.83\linewidth}{!}{%
			\begin{tabular}{cccccc}
				\hline
				\begin{tabular}[c]{@{}c@{}} Case \\number \end{tabular} &
				\begin{tabular}[c]{@{}c@{}}Total compliance\\ {[}$\frac{N.m}{rad}${]}\end{tabular} &
				\begin{tabular}[c]{@{}c@{}}$\lambda_{Passive}$\\ {[}\%{]}\end{tabular}  &
				\begin{tabular}[c]{@{}c@{}}Control frequency\\ {[}Hz{]}\end{tabular} &
				\begin{tabular}[c]{@{}c@{}}Delay\\ {[}ms{]}\end{tabular} \\ \hline
				1 & 1.6717 & 100 & 1000 & 0 \\
				2 & 1.6717 & 100 & 1000 & 0 \\
				3 & 1.6717 & 0 & 1000 & 0 \\
				4 & 1.6717 & 0 & 1000 & 17 \\
				5 & 2.5076 & 67 & 200 & 27 \\ 
				6 & 2.5076 & 67 & 100 & 27 \\
				7 & 2.8419 & 59 & 100 & 35 \\ \hline
			\end{tabular}
			}
		\caption{Simulation parameter of quadrupedal robot for task of landing with different control frequency and $\lambda_{Passive}$ }
		\label{tab:case}
	\end{table}
	
\end{document}